\documentclass[letterpaper,10pt,conference]{ieeeconf}
\IEEEoverridecommandlockouts
\usepackage{amsmath,amssymb}
\usepackage{booktabs}
\usepackage{cite}
\usepackage{graphicx}
\usepackage{overpic}
\usepackage{microtype}
\usepackage{url}
\usepackage{float}

\newcommand{\Nm}{\ensuremath{\mathrm{N}\!\cdot\!\mathrm{m}}}

\graphicspath{{figures/}}

\title{\LARGE \bf
LETS: Low-Overhead External-Torque Sensing for Force-Feedback
Bilateral Teleoperation}

\author{Zonghao Guo$^{1,2}$, Dongchen Guan$^{3}$, Yubo Wang$^{2,1}$, and Peng Yang$^{1,2}$%
\thanks{$^{1}$Institute of Automation, Chinese Academy of Sciences, China.}%
\thanks{$^{2}$School of Artificial Intelligence, University of Chinese Academy of Sciences, China.}%
\thanks{$^{3}$Beijing Amigos Robot Technology Co., Ltd., China.}%
\thanks{Corresponding author: Zonghao Guo (guozonghao2025@ia.ac.cn).}}

\begin{document}

\bstctlcite{BSTcontrol}

\maketitle
\thispagestyle{empty}
\pagestyle{empty}

\begin{abstract}
Bilateral teleoperation with force feedback is becoming increasingly important,
and the quality of external-torque estimation is critical to this process.
However, manufacturer-provided disturbance-torque estimates on commercial
robotic arms are often inaccurate, while the use of end-effector force/torque sensors adds cost
and payload. We propose LETS, a physics-guided learning approach that assigns
rigid-body dynamics to recursive Newton--Euler inverse dynamics and reserves
learning for only the robot-specific residual. LETS uses only the robot's URDF
model and joint feedback, deliberately excluding control commands and thereby
decoupling external-torque estimation from control-command inputs.
It trains on 12 minutes of free-motion
data in under one minute, without requiring a dedicated end-effector
six-axis force/torque sensor. In bilateral
teleoperation, LETS achieves real-time external-torque estimation that enables
stable force feedback. Its free-space error is 69\% lower than the
manufacturer-provided estimate, substantially reducing coupled
oscillations between leader and follower; its contact error is 28\% lower,
yielding more accurate haptic feedback.
\end{abstract}


\section{Introduction}

With the rapid development of embodied AI, acquiring high-quality robot
manipulation data via bilateral teleoperation has become increasingly important.
Force feedback is critical for this process: accurate external-torque estimation
enables transparent haptic rendering and contact-rich demonstration collection
\cite{pacchierotti2024haptic,liu2025factr,oh2026factr2}. Estimation errors
introduce spurious force sensations, attenuate true contact signals, and can
destabilize the coupled teleoperation loop \cite{lawrence1993stability}. Beyond
data collection, accurate force estimates serve as policy modalities for
contact-rich robot learning, making reliable external-torque sensing a
foundational capability for embodied intelligence.

External force is commonly obtained in one of two ways: direct measurement with
end-effector force/torque sensors or joint-level torque estimation. The former provides direct
measurement but adds cost, payload, and wiring complexity, and cannot detect
contacts occurring on upstream links or other parts of the robot
\cite{shan2024fine,oh2026factr2}. Joint-level estimation avoids
these limitations but faces accuracy challenges. Manufacturer-provided
disturbance-torque estimates are often too biased or noisy for transparent haptic rendering.
Model-based methods using nominal dynamics (e.g., recursive Newton--Euler
\cite{featherstone2008rigid})
omit transmission, friction, and actuator effects that dominate real-robot
residuals. Pure-learning approaches can capture these effects but must
first recover the rigid-body structure that analytical models provide for free.
This requires substantial training data and risks suppressing contact signals
when the learned predictor generalizes to unseen interactions. Recent work
addresses the data challenge by training on free-space motion
\cite{oh2026factr2,dou2026neuralactuator}, but the structure-recovery burden
remains.
Worse still, pure-learning estimators face an inherent dilemma in bilateral
teleoperation: if control commands are excluded, the network struggles to
reconstruct the rigid-body dynamics from finite data, yielding poor accuracy in
estimating internal joint torques; if command errors are included, the network
inevitably encodes the feedback gains into its weights---leading to contact leakage,
entanglement with control parameters, and excessive reliance on real-time,
reliable command signals. Both approaches suffer from structural flaws, making
it difficult to provide a robust solution for closed-loop haptic feedback.
A detailed analysis of this dilemma and our formulation to
resolve it is provided in Section III.
\begin{figure}[!t]
  \centering
  \includegraphics[width=\columnwidth]{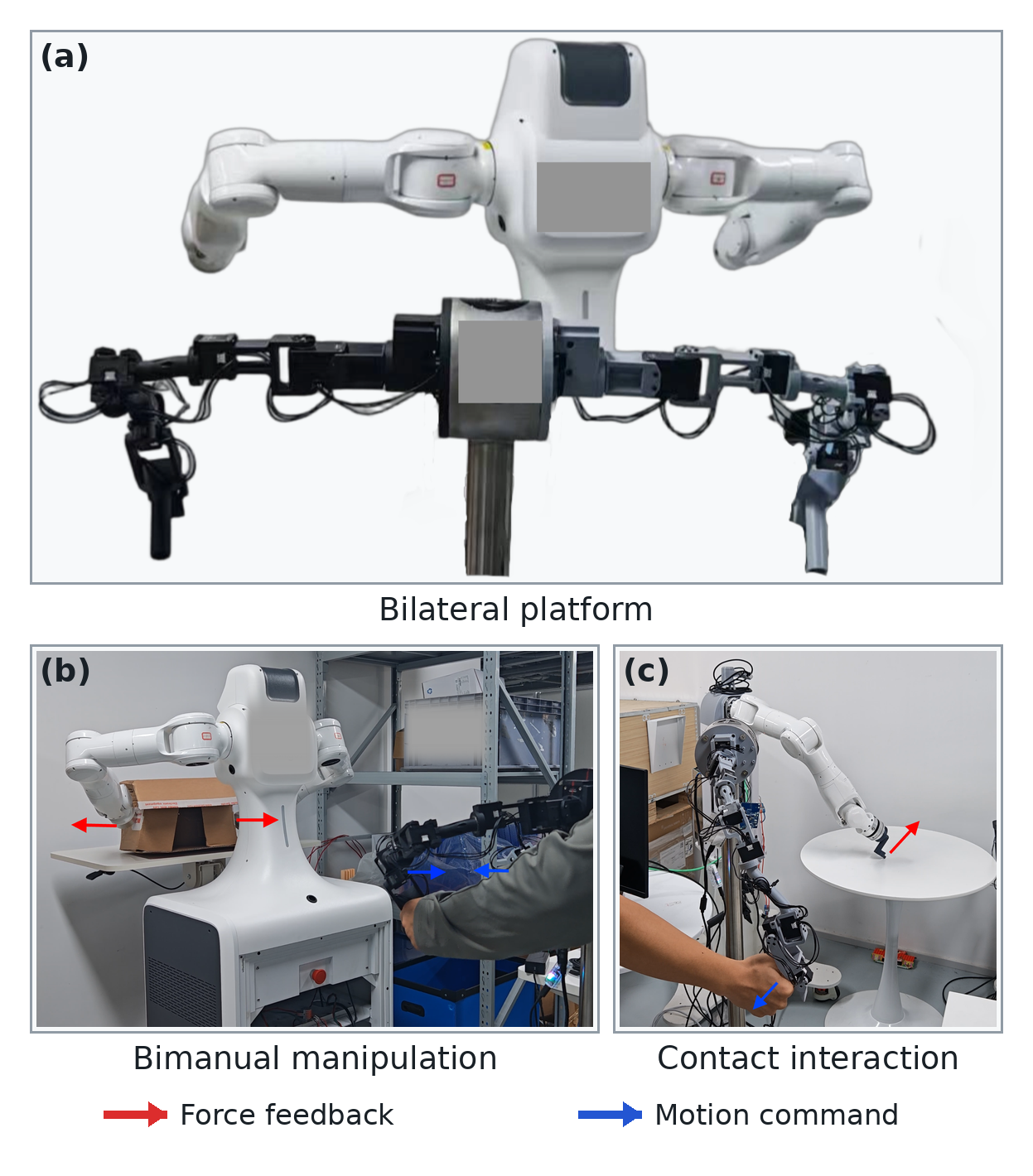}
  \caption{System overview: (a) bilateral platform, (b) bimanual box
  manipulation, and (c) single-arm contact interaction. Red arrows denote
  force feedback; blue arrows denote motion commands.}
  \label{fig:system}
\end{figure}

To address this, we propose LETS (\textbf{L}ow-overhead
\textbf{E}xternal-\textbf{T}orque \textbf{S}ensing), a physics-guided learning
method for external torque estimation. LETS decomposes joint torque during
free-space motion into two components: a nominal rigid-body dynamics component
calculated via the Recursive Newton--Euler Algorithm (RNEA), and a learned
temporal residual component that captures transmission, friction, and actuator
effects. This residual network relies solely on the history of joint states,
eliminating the need for command tracking. This approach reduces data
requirements, maintains sensitivity to contact signals, preserves the
analytical dynamic structure---which supports stable estimation during
contact---and supports model interpretability and safer interaction. By
providing stable, low-bias force feedback, the method helps operators avoid hard
collisions and excessive grasping forces that could damage manipulated objects.
Training labels are derived from measured torques during free motion; upon
deployment, LETS extracts the external interaction torque by subtracting the
learned free-space prediction from the measured torque. The model is trained
in 30 seconds on 12 minutes of free-motion data and operates at a deployment
frequency of 200~Hz. We deploy LETS as the force-sensing layer in
the bilateral teleoperation platform shown in Figure~\ref{fig:system}, where an
operator controls a follower unit via an actuated leader unit.

This paper makes three contributions:

\textbf{Low-overhead external-torque sensing from joint feedback alone.}
LETS combines fixed RNEA inverse dynamics with a lightweight causal temporal
residual learner. The network uses only joint-state history and deliberately
excludes all command signals, estimating external torque from the robot's URDF
and measured joint feedback without requiring a dedicated end-effector six-axis
F/T sensor at runtime. This design maintains high estimation performance without
external force hardware or controller-side signals, improving reusability and
reliability across teleoperation configurations.

\textbf{Data-efficient physics-guided learning.}
Explicit rigid-body factorization substantially reduces the data required for
robot-specific adaptation. With only 10\% of training windows, the physics-guided
variant outperforms its pure-learning counterpart trained on all windows; under
matched full-data training, the same factorization reduces error by 47\%.

\textbf{Controller-decoupled and robust bilateral force feedback.}
By using only robot-side joint feedback, LETS decouples external-torque sensing
from motion commands and controller-side signals. In real bilateral tests, LETS
maintains smooth free motion and stable rigid contact even when intermediate
command smoothing is bypassed, whereas baseline feedback exhibits pronounced
unloaded oscillations. Independent six-axis evaluation further confirms accurate
contact-torque preservation, supporting reliable force feedback across
teleoperation configurations.

\section{Related Work}

\subsection{External-Torque Estimation}

Acquiring reliable external-torque information faces a fundamental trade-off
between accuracy and practicality. End-effector force/torque sensors provide
direct measurement but add cost, payload, and cannot detect interactions away
from the tool center point \cite{shan2024fine,oh2026factr2}. Joint-level
estimation avoids these constraints but
faces persistent accuracy challenges.

Model-based approaches use inverse dynamics, momentum observers, and friction
identification \cite{deluca2005sensorless,kommuri2022external,bona2005friction,
roveda2021sensorless}. They provide interpretable structure but depend on
precise parameters and struggle with transmission nonlinearities, thermal drift,
and cable compliance. Pure-learning approaches can capture these unmodeled
effects \cite{haddadin2017learningcollision,callar2022hybridtimeseries,
long2023foeso}, but the learned model must
rediscover rigid-body dynamics from scratch.

Physics-guided learning addresses this by assigning rigid-body structure to
analytical models and learning only the residual
\cite{nguyentuong2010using,lutter2019deep,sutanto2020encoding,lim2024mobnet,yang2025hybrid,
zhou2026dualstage}. FACTR uses sensed external joint torque for bilateral
teleoperation and force-aware policy learning, rather than estimating it
\cite{liu2025factr}. NEXT, the torque estimator introduced in FACTR~2, learns
the entire free-space torque purely from joint-state history and command-position
error, without retaining a fixed physical model and therefore coupling
estimation to controller signals \cite{oh2026factr2}. NeuralActuator targets
broader actuator modeling using command and actuator telemetry in differentiable
simulation \cite{dou2026neuralactuator}. Yilmaz et al. \cite{yilmaz2024sensorless}
use an end-to-end LSTM to learn the dVRK's full free-space inverse dynamics,
whereas LETS uses URDF-derived rigid-body dynamics as a fixed prior and learns
only the robot-specific residual, avoiding full-dynamics identification and
making transfer substantially simpler. This factorization enables
controller-decoupled external-torque sensing for closed-loop force feedback.

\subsection{Force Information for Embodied Intelligence}

Embodied AI increasingly relies on contact-rich manipulation data. High-precision
grasping and assembly require reliable sensing and careful handling of contact
constraints \cite{li2019survey}. Force signals guide operators during
teleoperated collection, annotate contact phases for policy learning, and serve
as policy inputs
\cite{liu2025factr,oh2026factr2,suomalainen2022contact,zhao2023learning,
chi2024universal}. Obtaining reliable force information at scale, however,
remains challenging: dedicated sensors add cost that scales poorly across
data-collection fleets, while manufacturer-provided disturbance-torque estimates are often too
biased or noisy for transparent haptic rendering or reliable contact annotation.
Learning-based systems have explored force-aware policy representations and
contact-rich demonstration collection, but force acquisition typically relies on
either dedicated sensors or manufacturer-provided estimates with known limitations.

\subsection{Force Feedback in Bilateral Teleoperation}

Bilateral teleoperation with force feedback enables remote manipulation and
demonstration collection \cite{pacchierotti2024haptic}. Stability under delay and uncertain impedance has been
studied through control architectures such as four-channel bilateral control,
wave variables, and time-domain passivity
\cite{lawrence1993stability,anderson1989bilateral,niemeyer1991stable,
hashtrudi1993design}. These strategies assume reliable force feedback. In
practice, end-effector sensors add cost and payload, while manufacturer-provided
disturbance-torque estimates can introduce bias and noise that cause false sensations or
coupled oscillations. Learned estimators have improved estimation accuracy and
contact detection \cite{liu2021sensorless}, but validation has focused on
open-loop metrics rather than closed-loop bilateral feedback under operator
control.

\begin{figure*}[!t]
  \centering
  \includegraphics[width=0.98\textwidth]{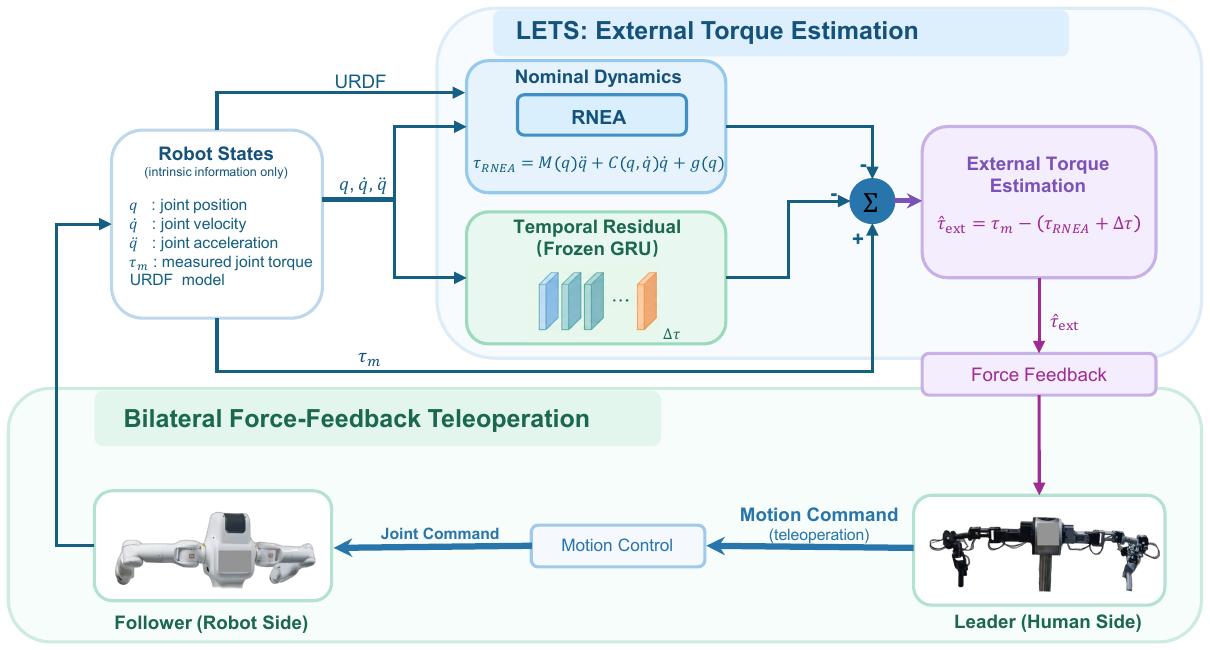}
  \caption{LETS online system architecture. Fixed RNEA uses the URDF and current
  joint state, while the frozen GRU uses joint-state history. Their predictions
  are subtracted from measured joint torque to estimate external torque at
  200~Hz. The estimate is mapped and scaled for leader-side force feedback;
  leader motion commands drive the follower, using joint impedance control in
  our experiments. Control commands do not enter the LETS estimator.}
  \label{fig:pipeline}
\end{figure*}

\section{Problem Formulation and Method}

\subsection{Problem Formulation}

The measured joint torque $\tau_m \in \mathbb{R}^7$ (read from the robot's
torque sensor) comprises free-space dynamics and external interaction:
\begin{equation}
\tau_m = \tau_{\mathrm{free}} + \tau_{\mathrm{ext}},
\end{equation}
where $\tau_{\mathrm{free}}$ includes gravity, inertia, Coriolis effects,
friction, and actuator dynamics, and $\tau_{\mathrm{ext}} \in \mathbb{R}^7$ is
the external torque applied by the environment (e.g., contact forces). Our goal
is to estimate $\tau_{\mathrm{ext}}$ causally from joint-level feedback:
\begin{equation}
\hat\tau_{\mathrm{ext}} = \tau_m - \hat\tau_{\mathrm{free}}.
\end{equation}

External joint torque $\tau_{\mathrm{ext}}$ is not directly observable and must
be inferred by subtracting the estimated internal dynamics from the total
measured torque. Robots with joint torque sensors can directly measure $\tau_m$;
those without require motor-current-to-torque calibration, whose quality
directly affects estimation accuracy. Calibration methods are outside the scope
of this work. LETS addresses a complementary challenge: given reliable total
joint torque, how can we estimate external torque in a low-cost, general way
that improves bilateral teleoperation and provides trustworthy force signals for
other system components?

\subsection{LETS Estimator}

LETS decomposes free-space torque into nominal rigid-body dynamics computed via
RNEA and a learned temporal residual. Offline training on free-motion data
produces the residual model. Online deployment (Fig.~\ref{fig:pipeline})
subtracts the learned free-space prediction from measured torque,
yielding external torque that is fed back to the leader for haptic rendering.

We use the Recursive Newton-Euler Algorithm (RNEA) \cite{featherstone2008rigid}
to compute nominal inverse dynamics:
\begin{equation}
\tau_{\mathrm{rnea}} = M(q)\ddot{q} + C(q,\dot{q})\dot{q} + g(q).
\end{equation}
Here, $M(q)$ is the inertia matrix, $C(q,\dot{q})$ captures Coriolis effects,
and $g(q)$ is gravity. The robot's factory URDF model
provides link parameters without modification. RNEA captures gravity, inertia,
and velocity coupling at their correct physical scale, but omits friction,
compliance, actuator dynamics, and parameter mismatch.

The free-space predictor combines RNEA with a learned residual:
\begin{equation}
\hat\tau_{\mathrm{free}} = \tau_{\mathrm{rnea}} + f_\theta(x_{1:T})
= \tau_{\mathrm{rnea}} + \Delta\tau.
\end{equation}
Here, $\Delta\tau = f_\theta(x_{1:T})$ denotes the predicted temporal residual
shown in Fig.~\ref{fig:pipeline}.
LETS does not prescribe a particular temporal network. For the manipulator and
data-collection protocol studied here, a validation-guided search over network
architectures and hyperparameters selected a single-layer 96-unit GRU for
$f_\theta$ with a 0.8-s input history ($T=160$ at 200~Hz). Each timestep's input is
$[\sin q_t, \cos q_t, \dot{q}_t, \ddot{q}_t] \in \mathbb{R}^{28}$. The network
is trained on twelve minutes of free-motion data to
predict the residual $y_t = \tau_{m,t} -
\tau_{\mathrm{rnea},t}$.
Crucially, we use state-only inputs and deliberately exclude all command signals
(e.g., desired positions or velocities). This design choice is two-fold: first,
it prevents contact leakage---command tracking error correlates with interaction
forces, and feeding it to the network would risk absorbing contact torques into
the internal-dynamics prediction. Second, and more fundamentally, it decouples
the residual estimate from the specific feedback controller, ensuring that the
learned model reflects only the robot's intrinsic actuation properties (friction,
transmission, compliance) rather than the control gains $(K_p,K_d)$. This
controller-agnostic property is essential for bilateral teleoperation, where
leader-side tuning and operator impedance vary across sessions. We test this
decoupling in Section IV.

At each 200~Hz control tick, the system computes RNEA, evaluates the GRU
residual after warm-up, and subtracts the learned free-space prediction from
measured torque to obtain $\hat\tau_{\mathrm{ext}}$.

The estimated torque is mapped to the leader's joint space and rendered as force
feedback. The manufacturer estimate and the LETS estimate use the same
downstream processing, so only the estimation source differs across conditions.

\section{Experiments}

The experiments address six questions:
\begin{enumerate}
  \item Does LETS outperform relevant baselines in free space and during contact?
  \item Does each component of the LETS factorization contribute to its performance?
  \item How much does the physics-guided structure reduce the required training data?
  \item Does LETS remain robust when operating conditions differ from training?
  \item Can the same protocol transfer to a second physical unit of the same manipulator model?
  \item Does the estimated torque remain stable in a real bilateral loop when intermediate trajectory smoothing is removed?
\end{enumerate}

\textbf{Protocol overview.} Experiments use a TianJi Marvin M6 seven-axis
manipulator, with a second physical unit of the same model reserved for the
transfer test. A Kunwei KWR75B six-axis F/T sensor mounted at the end effector
provides independent contact references and is used only for evaluation, not
for training or deployment. Ten free-motion runs were collected; the first eight
provide 12 min of training data, and the final two are reserved for validation
and testing. Nine subsequent free-space runs and nine contact runs with an
independent end-effector six-axis F/T sensor provide frozen evaluation. A second
robot unit tests transfer. All primary estimators use the same 200~Hz causal
interface; the fixed NEXT reproduction follows its original 100~Hz protocol.
No test or deployment run enters model selection. The complete protocol
appears in Table~\ref{tab:protocol}.

\begin{table}[H]
\caption{Experimental protocols and their roles.}
\label{tab:protocol}
\centering
\footnotesize
\setlength{\tabcolsep}{4pt}
\begin{tabular}{@{}lcll@{}}
\toprule
Protocol & Runs & Reference & Role \\
\midrule
Train/validation/test & 8/1/1 & $\tau_{\mathrm{ext}}=0$ & fitting/selection/locked test \\
Subsequent free space & 9 & $\tau_{\mathrm{ext}}=0$ & frozen primary evaluation \\
F/T-referenced contact & 9 & $\tau_{\mathrm{FT}}$ & frozen primary evaluation \\
Second physical unit & 7/1/1 & $\tau_{\mathrm{ext}}=0$ & fixed-protocol replication \\
Compound shift & 10 poses & $\tau_{\mathrm{FT}}$ & robustness stress test \\
\bottomrule
\end{tabular}
\end{table}

\textbf{Baselines and metrics.} We compare against the manufacturer-provided
disturbance-torque estimate (hereafter, manufacturer estimate), nominal RNEA, and a
generalized-momentum observer (GMO),
a standard model-based baseline for sensorless external-torque estimation
\cite{deluca2005sensorless}. Because the observer gain trades response speed
against noise sensitivity, we use $K=10$~rad/s as a fixed moderate-bandwidth
operating point, corresponding to a nominal 0.1~s time constant. We also test a
train-calibrated GMO with ridge-corrected friction and bias. The same observer
gain is fixed across every condition without retuning on deployment or contact
data. We also compare a pure-learning baseline formed by removing the RNEA branch
from LETS and training the remaining network to predict total free-space joint
torque; its architecture and hyperparameters are selected by validation.

We evaluate estimation error using RMSE and MAE. For run $r$ with $T_r$ samples,
$\hat{\tau}_{r,t,j}$ denotes the estimated torque for joint $j$ at sample $t$,
and $\tau^{\mathrm{ref}}_{r,t,j}$ is the corresponding reference. The reference
is zero in free space and is $\tau_{\mathrm{FT}}$ for contact and compound-shift
evaluation. We compute
\begin{equation}
E_r^{\mathrm{RMSE}}=\frac{1}{7}\sum_{j=1}^{7}
\sqrt{\frac{1}{T_r}\sum_{t=1}^{T_r}
\left(\hat{\tau}_{r,t,j}-\tau^{\mathrm{ref}}_{r,t,j}\right)^2},
\end{equation}
and
\begin{equation}
E_r^{\mathrm{MAE}}=\frac{1}{7T_r}\sum_{j=1}^{7}\sum_{t=1}^{T_r}
\left|\hat{\tau}_{r,t,j}-\tau^{\mathrm{ref}}_{r,t,j}\right|.
\end{equation}
RMSE is the primary metric because it retains torque units and penalizes large
errors, which are especially undesirable in force feedback. The run-equal
primary score is $E_{\mathrm{run}}=(1/R)\sum_{r=1}^{R}E_r^{\mathrm{RMSE}}$,
so each deployment run contributes equally. For distribution-shift evaluation,
each pose-specific recording is treated as one run. MAE is reported only for the fixed
comparison with NEXT, the torque estimator introduced in FACTR~2
\cite{oh2026factr2}, following its average-L1 evaluation protocol. Results
therefore focus on run-equal RMSE, with MAE reported for the NEXT comparison.

\subsection{Baseline Comparison in Free Space and Contact}

This experiment tests whether LETS improves torque estimation under held-out
free-space and contact conditions. Ten paired free-space runs were collected with an
empty end effector. The first
eight runs form training; the ninth run is used for validation and the
tenth for testing. No
history window crosses a run boundary. The train/validation/test unit is a
complete run rather than a random window, preventing adjacent windows from the
same motion episode appearing in both optimization and evaluation.

\begin{figure}[t]
  \centering
  \begin{overpic}[width=0.97\columnwidth]{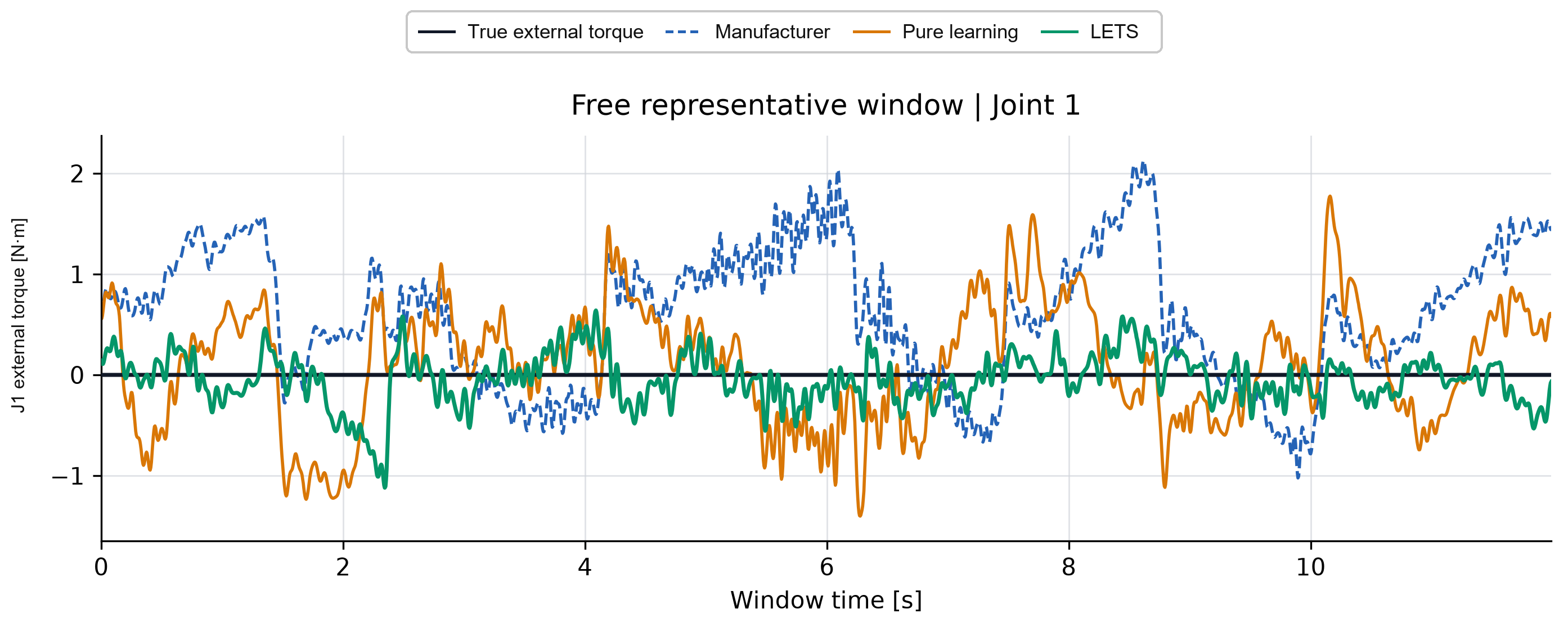}
    \put(1,37){\footnotesize\bfseries (a)}
  \end{overpic}\\[-2pt]
  \begin{overpic}[width=0.97\columnwidth]{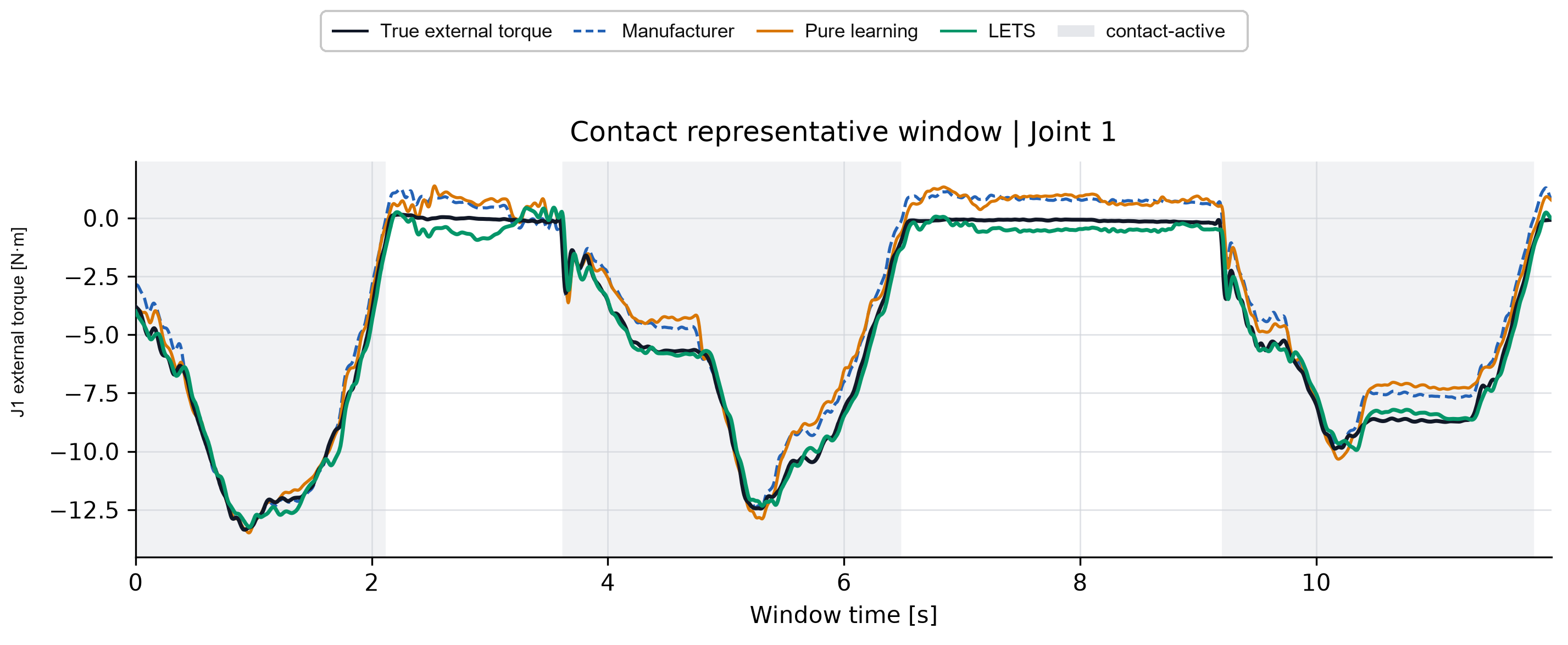}
    \put(1,37){\footnotesize\bfseries (b)}
  \end{overpic}
  \caption{Representative Joint~1 time-domain estimates from the frozen
  evaluation runs. (a) Free-space motion, where the external-torque reference is
  zero. (b) Rigid contact, where the reference is obtained independently from the
  six-axis F/T sensor as $\tau_{\mathrm{FT}}$.}
  \label{fig:j1_timeseries}
\end{figure}

For the nine contact evaluation runs, the independent end-effector sensor
measures the six-dimensional wrench $W_{\mathrm{FT}}$. Using the manipulator
Jacobian $J(q)$ computed from the robot URDF, we map this wrench to joint space
as $\tau_{\mathrm{FT}}=-J(q)^{\mathsf T}W_{\mathrm{FT}}$. This joint-space torque
serves only as the independent contact reference and is not used for training,
model selection, or deployment. Thus, contact errors in Table~\ref{tab:leaderboard}
and Fig.~\ref{fig:j1_timeseries} compare the estimated joint torque with
$\tau_{\mathrm{FT}}$, rather than with the raw wrench $W_{\mathrm{FT}}$; their
units are therefore $\Nm{}$.

\begin{figure}[t]
  \centering
  \includegraphics[width=0.90\columnwidth]{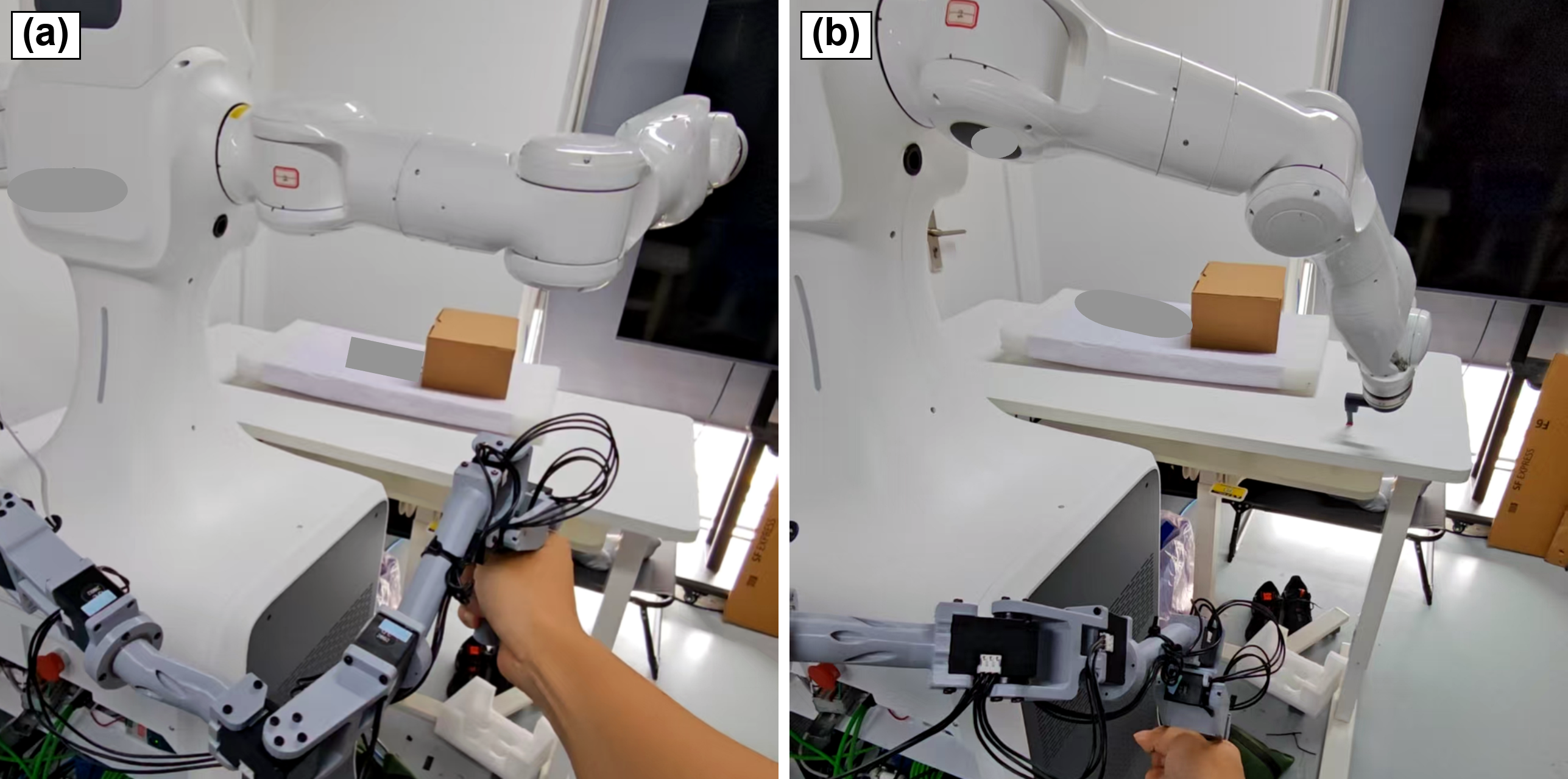}
  \caption{Feedback-sensitive conditions: (a) free-space bilateral operation
  and (b) rigid contact with an independently measured end-effector wrench.}
  \label{fig:conditions}
\end{figure}

Fig.~\ref{fig:conditions} shows the two feedback-sensitive operating conditions.
In free space, false torque appears as spurious operator effort and can excite
configuration-dependent coupled motion. During rigid contact, estimation noise
and delay can be returned as force variation and repeated rebound.

Table~\ref{tab:leaderboard} reports the run-equal mean per-joint RMSE for six
frozen estimators in the two conditions. In free space the reference is
$\tau_{\mathrm{ext}}=0$; in contact the reference is the independently measured
$\tau_{\mathrm{FT}}$ described above. Each condition contains nine evaluation runs.

\begin{table}[H]
\caption{Run-equal joint-torque RMSE (\Nm{}) over nine frozen runs.}
\label{tab:leaderboard}
\centering
\footnotesize
\begin{tabular*}{0.92\columnwidth}{@{\extracolsep{\fill}}lrr@{}}
\toprule
Estimator & \multicolumn{1}{c}{\shortstack{Free\\space}} &
\multicolumn{1}{c}{\shortstack{Contact\\($\tau_{\mathrm{FT}}$)}} \\
\midrule
Manufacturer estimate & 0.54 & 0.57 \\
Nominal RNEA & 0.53 & 0.55 \\
Nominal GMO & 0.51 & 0.69 \\
Calibrated GMO & 0.17 & 0.57 \\
Pure learning & 0.35 & 0.54 \\
\textbf{LETS} & \textbf{0.17} & \textbf{0.41} \\
\bottomrule
\end{tabular*}
\end{table}

\textbf{Free-space accuracy.} On nine frozen free-space runs collected after
training, LETS achieves 0.17~\Nm{} run-equal error versus 0.54~\Nm{} for the
manufacturer estimate (69\% reduction). The mean paired reduction is 0.38~\Nm{}, and LETS
is better in all nine runs.

\textbf{Contact preservation.} Low free-space error is not obtained by
suppressing contact. Against the independent joint-space reference
$\tau_{\mathrm{FT}}$ on nine separate contact runs, LETS achieves 0.41~\Nm{}
versus 0.57~\Nm{} for the manufacturer estimate (28\%
reduction), with improvement in all 9/9 runs.

Figure~\ref{fig:j1_timeseries} provides a time-domain view of the same
comparison. In free space, LETS remains closer to the zero-torque reference than
the manufacturer and pure-learning estimates. During contact, LETS follows the
independent $\tau_{\mathrm{FT}}$ reference with less bias across representative contact
transitions and sustained-contact intervals.

\textbf{Overall comparison.} LETS is the best contact estimator and matches the
best free-space error in this comparison. The nominal RNEA captures rigid-body
dynamics but omits robot-specific transmission and actuator effects, whereas the
pure-learning baseline must relearn the dominant rigid-body structure from data.
The two methods therefore each lack the component that describes the other part
of the robot dynamics. The nominal GMO performs reasonably in free space but
degrades markedly during contact, showing that a gain and calibration that fit
unloaded motion do not guarantee contact fidelity. The calibrated GMO approaches
LETS in free space but does not transfer cleanly to contact. Together, these
results support assigning known rigid-body dynamics to a fixed RNEA path and
learning only the robot-specific temporal residual.

\begin{figure*}[t]
  \centering
  \includegraphics[width=0.94\textwidth]{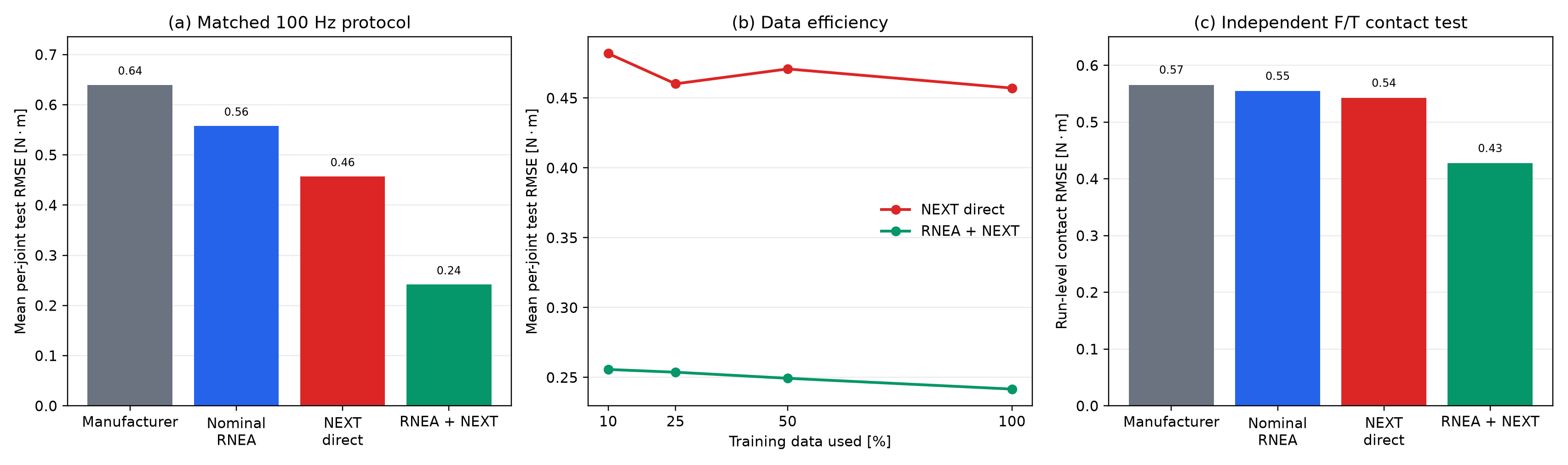}
  \caption{Controlled factorization study. (a) Mean per-joint test RMSE under a
  fixed NEXT architecture. (b) Uniform
  within-run subsampling demonstrates that 10\% of physics-guided
  training windows outperform all pure-learning windows. (c) Run-equal
  contact RMSE for validation-selected models over nine runs using the independent
  joint-space reference $\tau_{\mathrm{FT}}$.}
  \label{fig:controlled}
\end{figure*}

\subsection{Structural Ablation of Physics-Guided Learning}

Sec.~IV-A has already compared LETS with the pure-learning and nominal RNEA
baselines. We therefore use two controlled ablations here to isolate the torque
factorization and the deployed input choice.

\textbf{Controlled factorization under a fixed NEXT architecture.} We reproduce
NEXT using its reported 100~Hz, 0.5~s history, input, and network design
\cite{oh2026factr2}. Three seeds share normalization, windows, optimization,
and validation selection. Direct NEXT predicts measured free-space torque; the
RNEA+NEXT residual variant predicts the residual over RNEA. The two formulations are
matched in architecture, history, features, and optimization, changing only
the torque factorization.

Under the fixed NEXT architecture, explicit rigid-body factorization reduces
test error by 47\% (Table~\ref{tab:next}). This controlled factorization
ablation is separate from the deployed LETS architecture. Because the network, inputs, and
training settings are held fixed, this gain isolates the benefit of predicting
only the residual over RNEA rather than total torque. Panel (a) of
Fig.~\ref{fig:controlled} and Table~\ref{tab:next}
summarize the fixed-architecture comparison; panel (c) shows that the advantage
extends to independently referenced contact evaluation, where the RNEA+NEXT
residual variant achieves lower error across all nine contact runs.

\begin{table}[!htbp]
\caption{Fixed-architecture NEXT torque-estimation results (\Nm{}).}
\label{tab:next}
\centering
\footnotesize
\begin{tabular}{lrr}
\toprule
Method & RMSE & MAE \\
 & (\Nm{}) & (\Nm{}) \\
\midrule
Manufacturer raw output & 0.64 & 0.53 \\
Nominal RNEA & 0.56 & 0.48 \\
NEXT direct torque & 0.46 & 0.34 \\
RNEA + NEXT residual & \textbf{0.24} & \textbf{0.17} \\
\bottomrule
\end{tabular}
\end{table}

\textbf{Input ablation at the 200~Hz deployment protocol.} We additionally
compare LETS with an otherwise identical command-aware LETS
variant that appends the wrapped command-position error to its input history.
This comparison isolates the effect of adding command information while
keeping the LETS architecture and training protocol fixed.
Table~\ref{tab:input_ablation} reports run-equal mean per-joint RMSE in free space,
contact, and distribution shift; the last condition is analyzed in Sec.~IV-D.

\begin{table}[H]
\caption{LETS and command-aware input ablation at 200~Hz (RMSE in \Nm{}).}
\label{tab:input_ablation}
\centering
\footnotesize
\begin{tabular*}{0.96\columnwidth}{@{\extracolsep{\fill}}lrr@{}}
\toprule
Condition & LETS & Command-aware LETS \\
\midrule
Free space & \textbf{0.17} & 0.21 \\
Contact & \textbf{0.41} & 0.42 \\
Distribution shift & \textbf{0.52} & 0.59 \\
\bottomrule
\end{tabular*}
\end{table}

The command-aware endpoint is worse on all three held-out conditions. Thus,
command input provides no accuracy benefit in this matched comparison, while
the LETS design preserves controller independence and stronger
cross-condition robustness.

\subsection{Data Efficiency}

This experiment measures how much training data the physics-guided formulation
requires relative to pure learning. Reduced-data trials uniformly retain
windows within every training run, preserving workspace coverage while varying
supervision density.

Most notably, Fig.~\ref{fig:controlled}(b) shows that with only 10\% of training
windows, RNEA+NEXT outperforms direct NEXT trained on all windows (0.26 vs
0.46~\Nm{}). This demonstrates that assigning known physics to an analytical path
substantially reduces the data burden for robot-specific adaptation.

\begin{figure}[t]
  \centering
  \includegraphics[width=0.97\columnwidth]{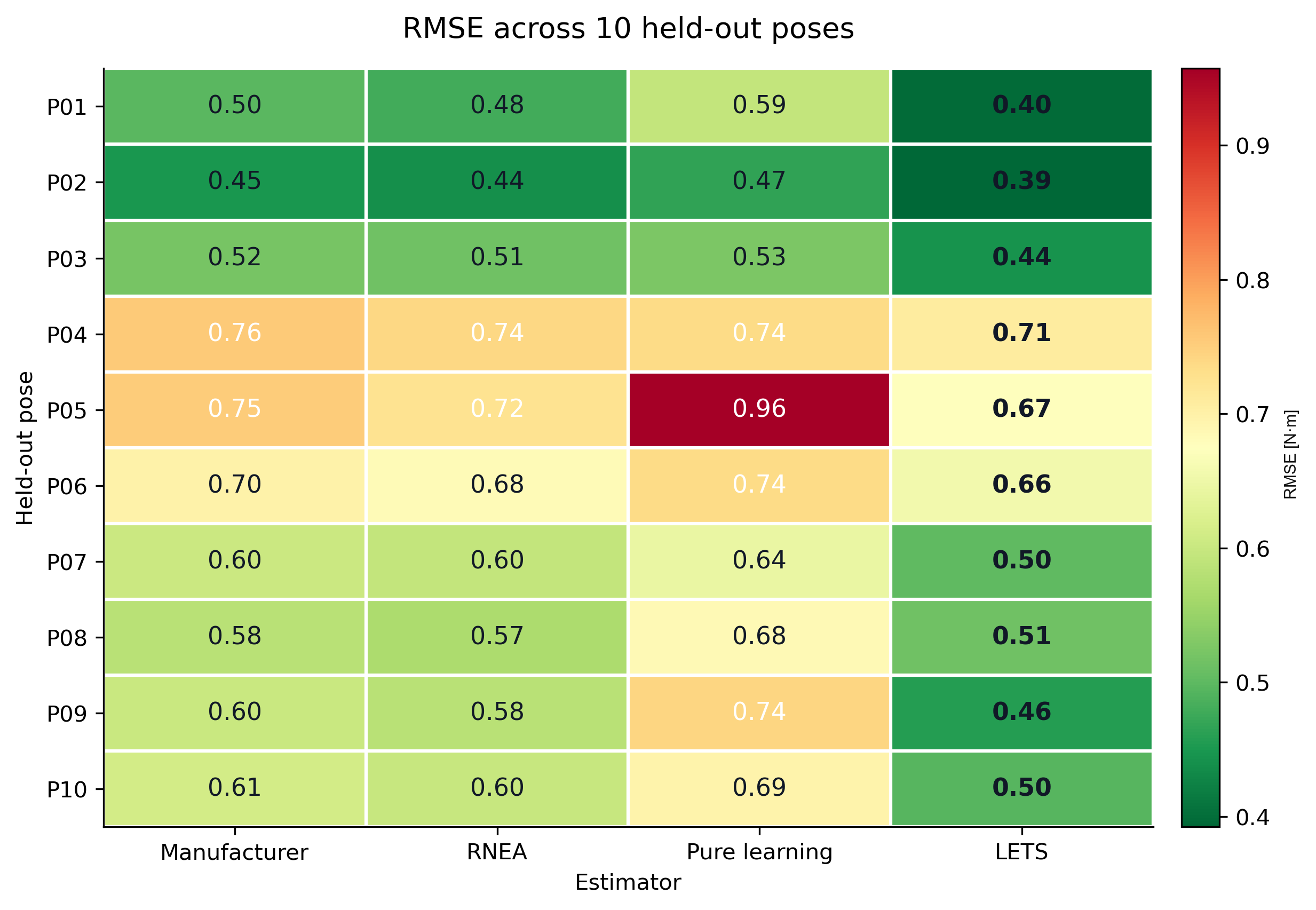}
  \caption{Compound-shift RMSE over ten held-out poses. Each cell reports the
  mean per-joint RMSE for one pose; lower is better. LETS is lowest in every
  pose-wise comparison.}
  \label{fig:ood_rmse}
\end{figure}

\subsection{Robustness Under Distribution Shift}

This experiment tests behavior beyond the training and bilateral-teleoperation
distributions. A compound-shift stress test averages active-contact RMSE equally
across ten position-specific recordings that change control context, motion
subspace, payload, and cable dynamics together.

The pure-learning baseline degrades to 0.68~\Nm{}, falling behind nominal RNEA
at 0.59~\Nm{} and the manufacturer estimate at 0.61~\Nm{}. In contrast, LETS
remains at 0.52~\Nm{}, below both model-based baselines
in all ten poses. Figure~\ref{fig:ood_rmse} shows the corresponding pose-wise
RMSE: LETS is the lowest-error estimator in all 10/10 out-of-distribution (OOD)
poses, while the
pure-learning estimator is more variable and reaches 0.96~\Nm{} at its worst
pose. This preservation of model-based robustness under compound shift
demonstrates that explicit rigid-body structure provides a stable foundation
beyond the training distribution.

\begin{figure*}[!t]
  \centering
  \includegraphics[width=0.98\textwidth]{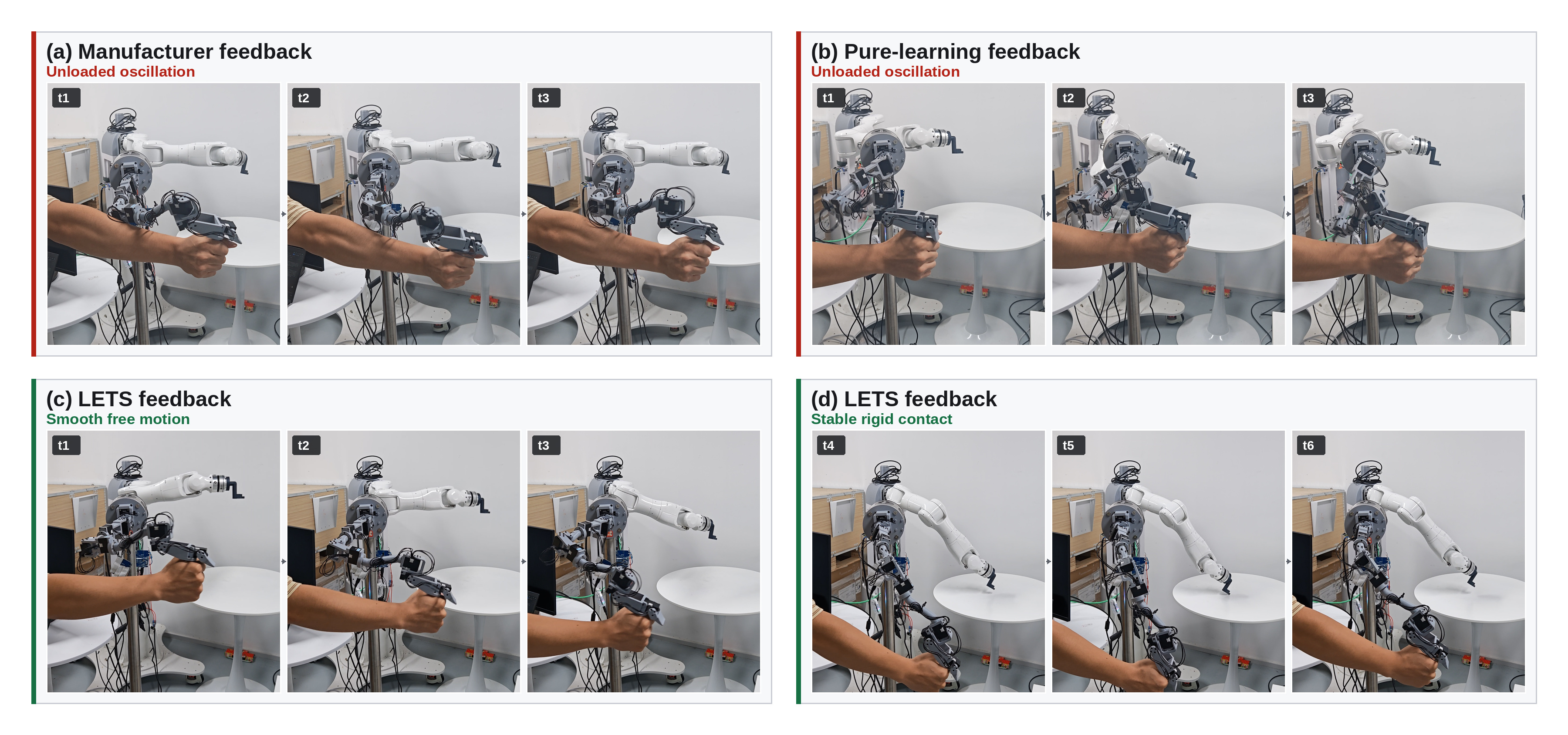}
  \caption{Real-robot closed-loop stress test with intermediate trajectory
  smoothing bypassed. Successive frames progress from left to right. Red denotes
  observed unloaded oscillation; green denotes smooth free motion or stable
  rigid contact. Continuous trajectories are provided in the supplementary
  video.}
  \label{fig:qualitative_loop}
\end{figure*}

\begingroup
\widowpenalty=10000
The fixed RNEA path provides the stable analytical backbone described above.
More importantly for distribution shifts, learning is restricted to the
smaller, repeatable robot-specific residual rather than the full torque signal,
reducing how much of the learned mapping must extrapolate beyond its training
distribution. The residual is also conditioned only on joint-state history and
does not encode the motion policy or controller commands that generated the
training data. This limits dependence on the original data-collection process
and explains the observed robustness under unseen operating conditions.
\par\endgroup

\subsection{Transfer to a Second Same-Model Manipulator}

This experiment tests whether the method transfers beyond a single physical
unit. The fixed training protocol was repeated on a second physical unit of the
same seven-axis model using 7/1/1 train/validation/test splits. The second unit
was installed at a different base pose rather than substituted at the first
unit's mounting position. Architecture, inputs, and optimization settings were
fixed; only unit-specific model parameters were fitted from the second unit's
free-space data.

LETS reduces held-out error from 0.71~\Nm{} (manufacturer estimate) and 0.66~\Nm{} (nominal
RNEA) to 0.19~\Nm{}, preserving the primary-unit ordering without another
architecture search. This result shows the value of decoupling external-torque
estimation from force rendering and bilateral control: the estimator is adapted
to the manipulator's own dynamics rather than to a particular leader, feedback
mapping, or installation. Once the LETS architecture and hyperparameters are
selected for a manipulator model, the same configuration can be rapidly adapted
to another physical unit by fitting only its unit-specific residual, without
repeating the model search.

\subsection{Real-Robot Closed-Loop Robustness}

In the standard bilateral loop used in the preceding experiments, leader joint
commands pass through Ruckig, an online trajectory-generation and smoothing
module, before being sent to the follower's servo-impedance controller. In this
stress test, that intermediate smoothing is bypassed and the leader commands
drive the follower controller directly. The resulting response is faster, with
sharper command and joint-torque variations, deliberately exposing
feedback-induced motion that trajectory smoothing could otherwise attenuate.

Figure~\ref{fig:qualitative_loop} shows the resulting closed-loop behavior.
With manufacturer-estimate feedback, large false external-torque estimates
immediately excite pronounced coupled oscillations between the leader and
follower. Pure-learning feedback performs better initially, but its
configuration-dependent estimation errors still produce large coupled
oscillations at some tested poses. Neither baseline therefore provides usable
feedback under this direct-feedback condition. In contrast, LETS remains smooth
across the tested poses and motions, without pronounced unloaded oscillation or
false resistance. During rigid contact, it returns a clear resisting response
without visible rebound or sustained oscillation.

These observations reflect the LETS factorization. Fixed RNEA provides a stable
physical backbone for the dominant robot dynamics, while the command-free
temporal residual models robot-specific effects from joint-state history without
directly introducing command transients into the estimate. LETS therefore
improves external-torque accuracy while keeping the estimator independent of
command smoothing and other controller-side signals, substantially increasing
closed-loop robustness under this demanding direct-feedback condition. The
complete continuous trials are provided in the supplementary video.

\section{Conclusion}

This paper addressed a practical gap in force-feedback bilateral teleoperation:
obtaining accurate external-torque sensing without a dedicated end-effector
six-axis F/T sensor at runtime. LETS assigns the dominant rigid-body dynamics to
a fixed RNEA path and learns only the robot-specific temporal residual from
causal joint-state history. This factorization keeps the estimator independent
of command signals while preserving a physically structured free-space torque
prediction.

The experiments show that this design provides more than lower nominal error.
Restricting learning to the residual substantially reduces the required
training data, while the fixed analytical path provides a stable foundation
when operating conditions differ from training. The command-free input further
limits dependence on the data-collection controller, allowing the same
architecture to adapt to a second physical unit and remain stable when
intermediate trajectory smoothing is removed. Independent six-axis contact
evaluation confirms that improved free-space accuracy does not suppress real
interaction torque, and the bilateral stress test connects that accuracy to
smooth free motion and stable contact in the signal experienced by the
operator. Together, these properties make LETS a reusable robot-side
force-perception layer rather than an estimator tied to one feedback controller
or teleoperation configuration.

The present evidence is limited to two physical units of one seven-axis
manipulator model and assumes a usable URDF model. In addition, the training-window
subsampling study does not yet determine how far the physical data-collection
duration or motion coverage can be reduced. Future work will extend the
adaptation protocol across robot morphologies, reduce the required collection
coverage, and evaluate broader operators, controllers, and contact-rich tasks.

\bibliographystyle{IEEEtran}
\bibliography{references}

\end{document}